\documentclass[letterpaper, 10 pt, conference]{ieeeconf}  

\IEEEoverridecommandlockouts                              

\usepackage[hidelinks]{hyperref}
\usepackage{lipsum} 
\usepackage{amssymb,graphicx,amsmath,color}
\usepackage{bm}
\usepackage{subcaption}
\usepackage{graphicx}
\usepackage{placeins}
\usepackage{gensymb}

\usepackage[percent]{overpic}
\usepackage{tabularx}
\usepackage{threeparttable}
\usepackage{courier} 
\usepackage{float}
\usepackage{algorithm}
\usepackage{algpseudocode}

\usepackage[noadjust]{cite} 

\usepackage{multirow}
\usepackage{rotating}
\usepackage{dblfloatfix}
\DeclareMathOperator{\atantwo}{atan2}

\title{\LARGE \bf
Shear-invariant Sliding Contact Perception with a Soft Tactile Sensor
}
 
\author{Kirsty Aquilina,  David A. W. Barton and Nathan F. Lepora
\thanks{KA was supported by the EPSRC Centre for Doctoral Training in Future Autonomous and Robotic
	Systems (FARSCOPE). NL was supported in part by a Leadership Award from the Leverhulme Trust on `A biomimetic forebrain for robot touch' (RL-2016-39)}
\thanks{Data used in this paper is available at http://doi.org/c24w}
\thanks{The authors are with the Bristol Robotics Laboratory, and the Department of
	Engineering Mathematics, University of Bristol, Bristol, U.K.\newline
    Email: \{ka14187, david.barton, n.lepora\}@bristol.ac.uk%
    }%
}

\begin{document}

\maketitle
\thispagestyle{empty}
\pagestyle{empty}

\begin{abstract}

Manipulation tasks often require robots to be continuously in contact with an object. Therefore tactile perception systems need to handle continuous contact data. Shear deformation causes the tactile sensor to output path-dependent readings in contrast to discrete contact readings. As such, in some continuous-contact tasks, sliding can be regarded as a disturbance over the sensor signal. Here we present a shear-invariant perception method based on principal component analysis (PCA) which outputs the required information about the environment despite sliding motion. A compliant tactile sensor (the TacTip) is used to investigate continuous tactile contact. First, we evaluate the method offline using test data collected whilst the sensor slides over an edge. Then, the method is used within a contour-following task applied to 6 objects with varying curvatures; all contours are successfully traced. The method demonstrates generalisation capabilities and could underlie a more sophisticated controller for challenging manipulation or exploration tasks in unstructured~environments.

\end{abstract}


\section{INTRODUCTION}

\begin{figure}[t!]
	\vspace{0.7 em}
	\centering
\begin{overpic}[width=0.38\textwidth]{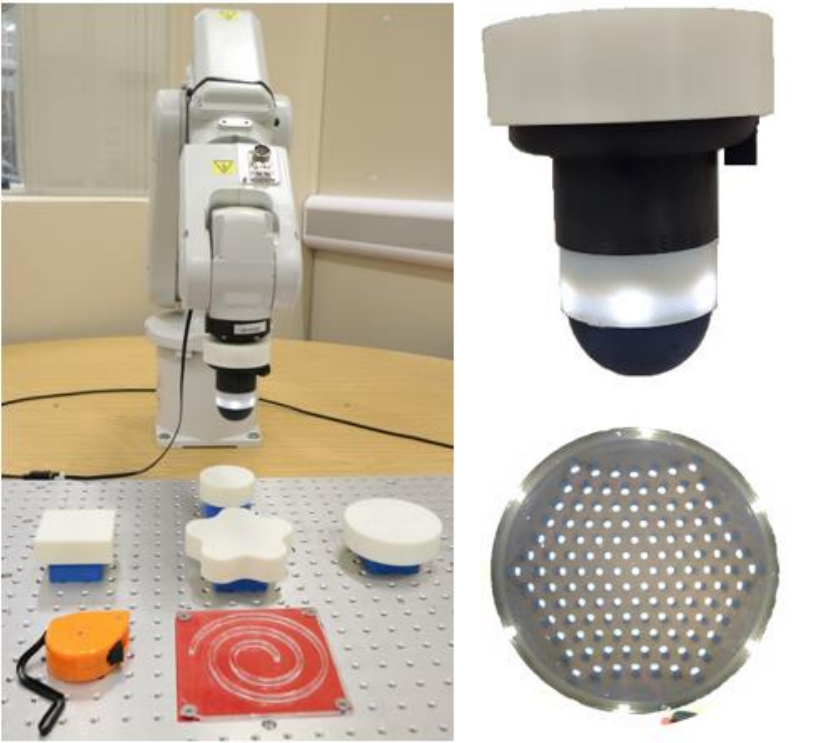}
\put(2,85){\textbf{(a)}}
\put(60,85){\textbf{(b)}}
\put(60,40){\textbf{(c)}}

\end{overpic}
	\caption{(a) The experimental setup showing four 3D-printed objects, a natural object (a tape measure) and a laser-cut spiral used for contour-following. The TacTip is the end effector of a 6 axis ABB robot (IRB120). (b) Close-up of the TacTip sensor. (c) The image captured by the TacTip at no contact.}
		\vspace{-1em}
	\label{fig:setup}
\end{figure}

Continuous contact sensing is crucial for robot manipulation or tactile exploration as these activities usually require the robot to be in continuous contact with objects. Furthermore, it is frequently desirable to use compliant touch sensors, which make the perception more challenging due to motion dependency caused by their sensitivity to shear \cite{Cramphorn2018}. Shear deforms the sensor (Fig.~\ref{fig:setup}) depending on the sliding direction, thus making sensor readings history dependent.

\begin{figure}
	\centering
	\begin{overpic}[scale=0.62]{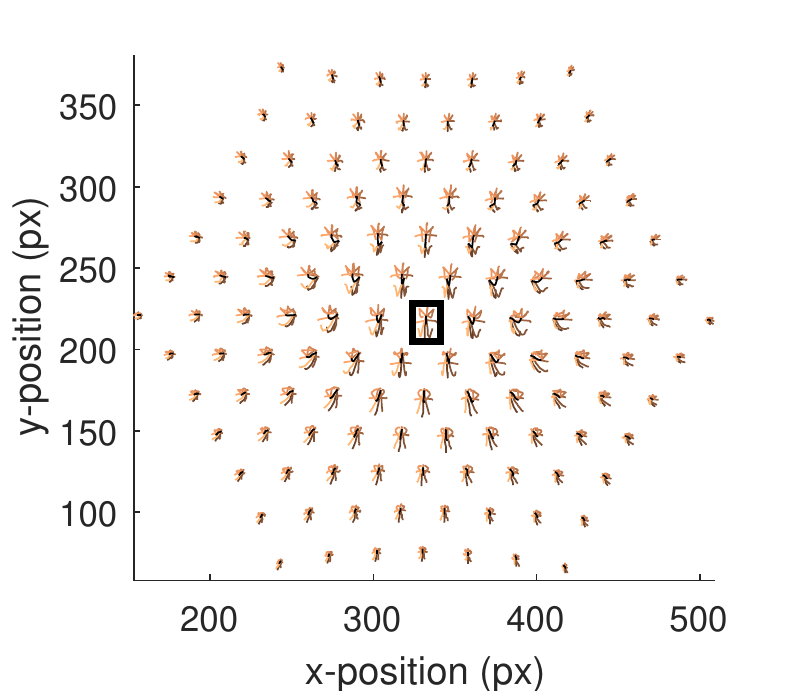}
		\put(-2,74.5){\textbf{(a)}}
		\put(86.5,74.5){\textbf{(b)}}
	\end{overpic}\hspace{-0.2cm} 
	\includegraphics[scale=0.62,trim={0 0 0 0.3cm},clip]{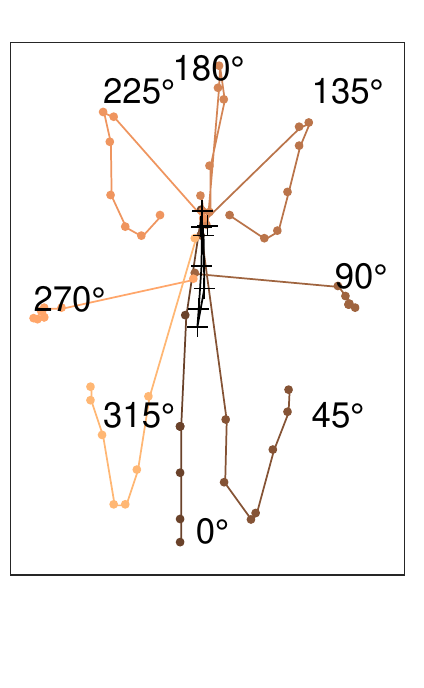}
	\caption{The effect of sliding motion on the tactile data. (a) The black trace shows the pin motion of the sensor when tapping across a straight edge at different lateral positions without sliding. The coloured traces depict the sensor moving the same amount across the edge as the black trace whilst sliding in different directions (from 0\degree{} to 315\degree{}). (b) Zoomed in version of (a) over the central pin marked by the central rectangle. Notice the difference between statically collected points (black) and those collected whilst sliding (coloured), showing that sliding influences the tactile sense.}
	\vspace{-1em}
	\label{fig:sliding}
\end{figure}

The novelty of this work is to verify the hypothesis that features found in independent tactile readings (taps) can also be extracted from data perturbed by sliding motion. Sliding motion causes the sensor skin to deform, and so sensor readings would depend on both the tactile features of the object and the shear direction. Thus, while discrete tap data is similar for the same tactile features, with sliding motion those features can produce completely different readings (Fig.~\ref{fig:sliding}). This paper finds a link between discrete tactile data (taps) and movement dependent data by showing that a linear transform can extract features of interest despite the sliding.

For validation, we apply a simple perception method trained on discrete contact data (taps) to continuous
contact data collected offline and to continuous
contact data for contour-following whilst sliding. The perception method uses principal component analysis (PCA) to extract features from
tactile data which are then mapped to the sensor pose using nonlinear regression. We also exploit PCA to visualise the
output of a soft biomimetic tactile sensor (the TacTip~\cite{chorley2009,Ward-Cherrier2018}, shown in Fig.~\ref{fig:setup}) and show that the data are strongly influenced by the sliding direction of the sensor (Fig.~\ref{fig:sliding}). This shear-invariant perception method extends previous tactile exploration studies \cite{Lepora2017,Martinez-Hernandez2017} that used tapping movements (independent tactile data) by demonstrating reliable sliding contact tactile exploration (contour-following) of various objects (Fig.~\ref{fig:setup}a) despite being trained using discrete contacts on a straight edge.

\section{BACKGROUND}

Sliding motion influences tactile sensing both in humans \cite{Srinivasan1990,Edin1995}, where the direction of motion results in different responses, and in artificial touch sensors, where shear deforms their outer soft layer. For example, sliding displaces the internal markers of the GelSight sensor \cite{Li2014,yuan2015} and introduces higher inaccuracies in a force/torque sensor embedded in a rubber-covered fingertip \cite{Liu2015}. Continuous tactile sensing during sliding can thus be a challenging task.

Tactile exploration is a commonly studied task which usually involves executing a sequence of discrete contacts on an object \cite{Yi2016a,Jamali2016,Martinez-Hernandez2017,Matsubara2017,Lepora2017}. Some have addressed this by using a Gaussian process (GP) model of the surface and exploring the most uncertain regions \cite{Yi2016a,Jamali2016}, or by combining the uncertainty with the cost of travel \cite{Matsubara2017}. Others used the perceived edge orientation to perform contour-following of a flat object \cite{Martinez-Hernandez2017,Lepora2017}.

Other studies perform continuous tactile exploration where the sensor is constantly in contact. Some of these studies perform tactile exploration by using a tactile servoing approach \cite{Li2013,Kappassov2016}, whilst others explore along the tangential plane of the surface \cite{Liu2015,Back2014,Driess2017}.  Typically these use either flat tactile sensor arrays \cite{Li2013,Kappassov2016}, which provide limited tactile feedback through tactile images, or force-torque measurements only \cite{Liu2015,Back2014,Driess2017}. There is thus a need for more studies that address continuous tactile exploration using tactile sensors with complex outputs (such as biomimetic sensors) that provide more information but that at the same time require a more complex perception system. In this work we show that sliding contact tactile exploration can be achieved using a complex touch sensor with an adequate perception system.

Previous work with the TacTip mostly involved discrete contacts (taps) with a surface for localization \cite{Lepora2015}, tactile exploration \cite{Lepora2017} or performed a rolling motion \cite{Cramphorn2016,Ward-Cherrier2017} or slip detection \cite{James2018}. This study takes a different approach from \cite{Lepora2015,Cramphorn2016,Ward-Cherrier2017,Lepora2017} by using unsupervised and supervised learning to perform sliding motion over an edge stimulus.

\section{METHODS} 

\subsection{Experimental Setup}
\label{sec:setup}

\subsubsection{Hardware}
The setup consists of a 6 axis robot arm (ABB IRB120) with a biomimetic optical tactile sensor (the TacTip~\cite{chorley2009}) as an end effector (Fig.~\ref{fig:setup}a). The TacTip is a multi-material 3D-printed sensor with a 40\,mm rubber-like dome with 127 white internal pins placed in a hexagonal arrangement (Figs.~\ref{fig:setup}b, c). The dome is filled with a gel that makes it soft and compliant. A USB camera tracks each individual pin in each recorded frame using image processing techniques \cite{Ward-Cherrier2018}, thus the sensor outputs the 2D pixel positions of each pin for each frame; refer to \cite{Ward-Cherrier2018} for further details.

The stimuli used in the experiment are four 3D-printed plastic objects, one natural object (a tape measure) and one laser-cut acrylic spiral~\cite{Lepora2017}. The 3D-printed shapes are a rectangle, two differently sized circles and a 5-petal flower-like shape. The objects are rigidly fixed in front of the robot.
\subsubsection{Training set experimental procedure} \label{subsubsec:trainingProcedure}
We collected a training set by tapping the sensor over the rectangular stimulus edge at different lateral positions, depths and orientations (Fig.~\ref{fig:setup}a). During a tap, the robot reaches the required sensor lateral position and orientation and then moves down onto the object to the required depth. The collected dataset comprises 288 poses (18 orientations $\times$ 8 lateral positions $\times$ 2 depths), with 1 frame recorded at each sensor pose.

The sensor rotates in 20\degree{} increments from -160\degree{} to 180\degree{}. At each sensor orientation $\theta_i$ ($i\in\{1,\dots,18\}$), the sensor moves to 8 different lateral positions $l_i$ with $i\in\{1,\dots,8\}$ (Fig.~\ref{fig:training}); namely, 9.9\,mm and 6\,mm to -6\,mm in \mbox{-2\,mm} steps. The middle of the sensor is in contact with the edge of the object at the 0\,mm position, while at the 9.9\,mm position the sensor is not touching anything. At each lateral position, the sensor moves to two different depths:~2\,mm and 4\,mm into the object. Data were collected at different depths to ensure the perception method would still generalise in~case the stimuli are not perfectly aligned. The perception label $w_i$ refers to either the $l_i$ label or $\theta_i$ label.

\begin{table}[t]
	
	\centering
	
	\setlength{\tabcolsep}{0.1em}

	\begin{tabular}{ccc}
		
		\includegraphics[grid,width = 0.25\textwidth]{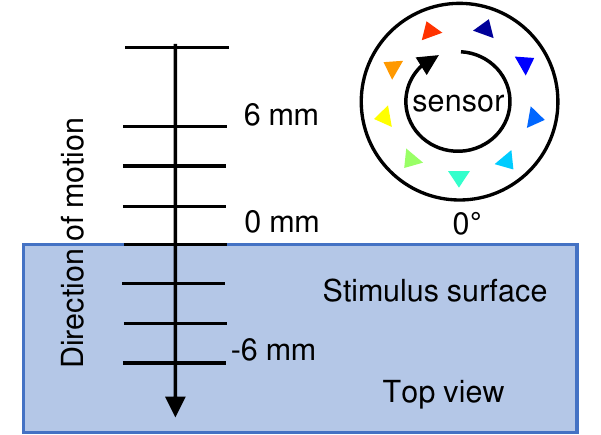}&
		\includegraphics[width =0.05\textwidth,trim={1.2cm 1.1cm 0 0},clip]{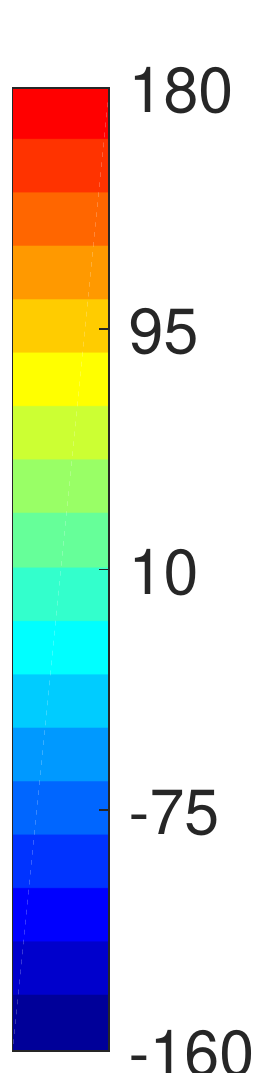}&

		\begin{sideways}{\textsf{ \small \hspace{0.25cm} orientation (degrees)}}\end{sideways}	
	\end{tabular}
	\captionsetup{type = figure}
	\caption{Training performed on a straight edge. At each orientation the sensor moves forward along the arrow without touching the object until it reaches the required position, then the sensor descends to touch the object (tap). The horizontal lines show the lateral training positions. Position 0\,mm aligns the middle of the sensor with the edge.}
	\vspace{-1.2em}
	\label{fig:training}
\end{table}

\subsection{Algorithms}

\subsubsection{Data Representation} \label{subsection:dataRep}

The data recorded by the TacTip consists of the $x$ and $y$ pixel positions of the 127 internal pins, as observed in each camera frame. The data is a column vector of these positions:
\begin{equation}
\label{multi dim data}
\bm{z} =[s_1, \dots, s_{N_{\rm dims}}]^T
\end{equation}
where $s_k$ is a pixel position value, $k$ is the dimension index and $N_{\rm dims}$ is the total number of sensor dimensions (here 254). The training data is stored as a data matrix, $D =\setlength\arraycolsep{1pt}
\begin{bmatrix}\bm{z_1},&\dots, &\bm{z_N}\end{bmatrix}^T$, where $N$ is the total number of collected frames during the experiment with $N =$ 288 (the total number of sensor poses).

\subsubsection{Principal Component Analysis and Data Transformation}

\label{sec:methodsPCA}

Aquilina et al. \cite{Aquilina2018} showed that when the taxels (sensing elements) of a touch sensor are correlated to each other, linear dimension reduction (PCA) can uncover useful structure in the data. This applies to the TacTip \cite{Aquilina2018}, which motivates the use of PCA in this work and where we extend it by showing that features found in tap data can also be used for sliding contact data.

PCA is performed on the data matrix $D$ and the first 3 principal components (PCs), are used in the rest of this study. These PCs, which are the projections of $D$, cumulatively describe 88.3\% of the signal variance in $D$. Each frame $\bm{z}$ is projected into vector $\bm{p}$ using \mbox{$\bm{p}\!=\!\setlength\arraycolsep{2pt}\begin{pmatrix} 
 \bm{e_1} & \bm{e_2} & \bm{e_3}
 \end{pmatrix}^T(\bm{z}-\bar{\bm{z}})$}
where $\bm{p}$  is a 3D column vector, $\bar{\bm{z}}$ is the mean of each column in $D$, and $\bm{e_1}$, $\bm{e_2}$ and $\bm{e_3}$ are the column eigenvectors.

Motivated by the correlation between sensor orientation and the angle of $\bm{p}$ in the 2\textsuperscript{nd} and 3\textsuperscript{rd}~PC plane \cite{Aquilina2018}, we transform each vector $\bm{p}$ into modified spherical coordinates $\bm{v}$. The vector $\bm{v}$ consists of $\rho$, $\theta_{\text{PC23}}$ and $\phi$ where \mbox{$\rho^2\!=\!\text{PC1}^2\!+\!\text{PC2}^2\!+\!\text{PC3}^2$}, \mbox{$\theta_{\text{PC23}}\!=\!\atantwo(\text{PC3},\text{PC2})$} and \mbox{$\phi\!=\! \atantwo((\text{PC2}^2\!+\!\text{PC3}^2)^{\frac{1}{2}},\text{PC1})$}. The operator $\atantwo$ is the four quadrant inverse tangent operator which considers the full 360\degree{} range and PCN is the N\textsuperscript{th} PC. The first no contact point collected is set as the origin, unless there are multiple depths in the training set in which case the origin used to compute $\rho$ and $\phi$ is shifted in the PC2-PC3 plane for each $\theta_{PC23}$ so that the same lateral positions in the middle of the collected range for different depths have the same $\phi$.

\begin{figure}[t!]

	\vspace{0.175cm}
	\centering
	\includegraphics[width = 0.55\linewidth]{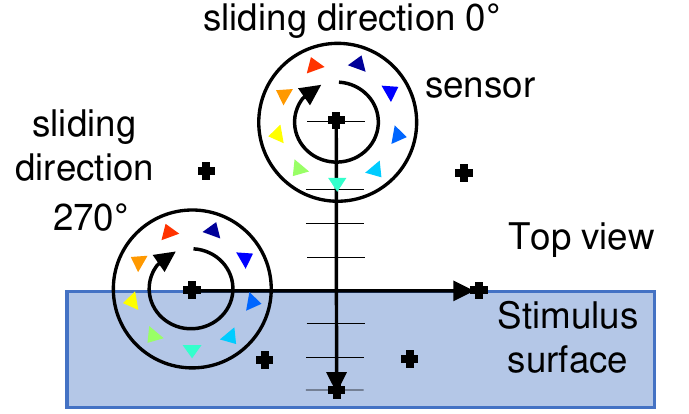}
	\caption{Multi-directional data collection procedure. The sensor slides along different directions at 45\degree{} increments. Along each direction the sensor pose varies as seen in Fig.~\ref{fig:training}, so the same colour bar applies here. The cross markers indicate the starting position of the sensor's midpoint and the arrow indicates the sliding direction. Here a 90\degree{} motion means that the sensor slides westward along the edge.}
	\label{fig:omnidirection}
	\vspace{-1.1em}
\end{figure}

\subsubsection{Training Data Pruning}

Data pruning is the elimination of noisy or mislabelled data from the training set to improve the performance of the learning algorithm \cite{Hong2008,Angelova2005,Kubica2002}.
Points collected at no contact during the training procedure have different physical quantity labels but are in reality indistinguishable. These points would introduce a discontinuity in a regression-based algorithm, therefore we remove them from the training set.

We use the sensitivity measure $S$ introduced in Aquilina et al. \cite{Aquilina2018} to remove the indistinguishable data points. The sensitivity is defined by 
\begin{equation}
\label{eq:sens}
 S = \frac{||\Delta\bm{p}||}{|\Delta w_i|}
\end{equation}
where $||\Delta\bm{p}||$ is the change in the projected vector $\bm{p}$ and $|\Delta w_i|$ is the change in label values (either orientation or lateral position). A large sensitivity  $S$ denotes a large change in sensor measurements $||\Delta\bm{p}||$ for a fixed change in the physical quantity of interest $|\Delta w_i|$, thus poses with a small $S$ are the most difficult to perceive. Using the sensitivity measure $S$ of each training point we eliminate outlying data from the training set by comparing it to a decision criteria. Following Hoaglin et al. \cite{Hoaglin1986}, the Tukey outlier detection rule (boxplot) is used to remove outliers that have a large $S^{-1}$, thus keeping only the easily distinguishable observations. An upper limit is used for the lateral position training set to ensure that at least 90\% of the data is always kept.

\begin{figure}[t!]

		\vspace{0.175cm}
	\centering
	\includegraphics[width = 0.48\textwidth]{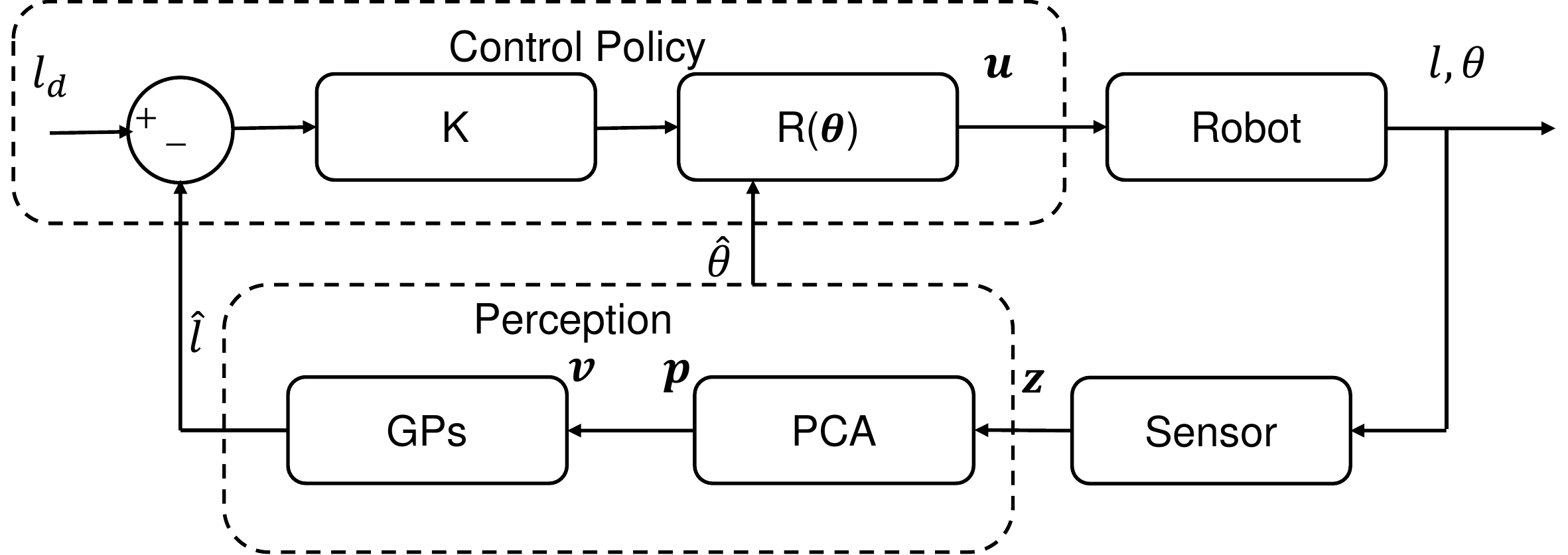}
	\caption{Flowchart of the overall contour-following system. The perception block consists of a PCA module which projects the sensor data into a 3D vector $\bm{p}$, this is transformed into spherical coordinates $\bm{v}$  which are the input of the GPs that output $\hat{\theta}$ and $\hat{l}$. The control policy block shows the dependencies of the control variable $\bm{u}$ on the estimated orientation $\hat{\theta}$ and the P controller used to reach a desired lateral position $l_d$.}
	\label{fig:contourMethod}
	\vspace{-1.2em}
\end{figure}

\begin{table*}[t]
		\vspace{0.15cm}
	\centering

	 \sffamily
		\begin{tabular}{ccccc}

			& {\normalsize \textbf{Sliding Direction 0\degree{}}} & {\normalsize  \textbf{Sliding Direction 90\degree{}}} & {\normalsize  \textbf{Sliding Direction 180\degree{}}} & {\normalsize  \textbf{Sliding Direction 270\degree{}}}\\
			
			\raisebox{0.5\totalheight}{\rotatebox[origin = lt]{90}{\mbox{\small Second PC} }}
			
			& 	\includegraphics[width = 0.19\textwidth, trim={0 0 0 0.1cm},clip]{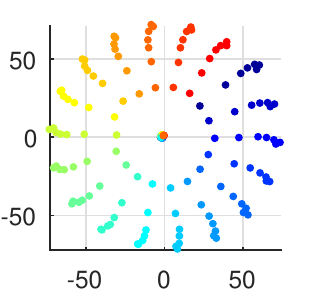}
			& 	\includegraphics[width = 	0.19\textwidth, trim={0 0 0 0.1cm},clip]{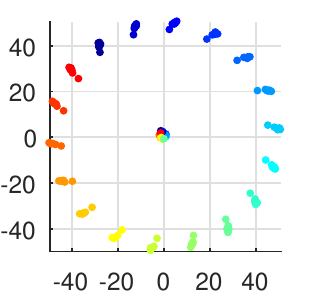}
			& 	\includegraphics[width = 0.19\textwidth ,  trim={0 0 0 0.1cm},clip]{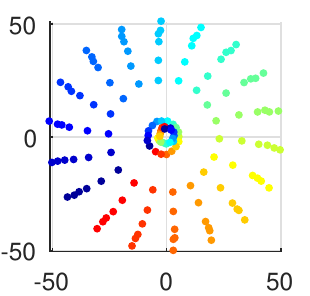}
			& 	\includegraphics[width = 0.19\textwidth,  trim={0 0 0 0.1cm},clip]{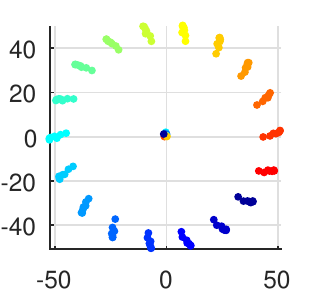}\\
			& {\small First PC} 
			& {\small First PC}
			& {\small First PC}
			& {\small First PC}\\

			\raisebox{0.5\totalheight}{\rotatebox[origin = lt]{90}{\mbox{\small Fifth PC} }}

			& \includegraphics[width = 0.19\textwidth, trim={0 0 0 0.1cm},clip]{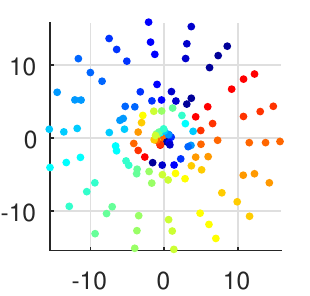}
			& \includegraphics[width = 0.19\textwidth, trim={0 0 0 0.1cm},clip]{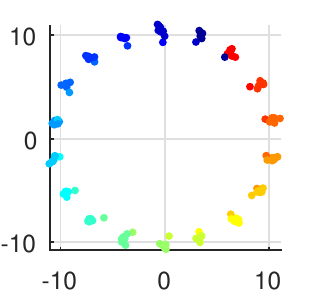}
			& \includegraphics[width = 0.19\textwidth, trim={0 0 0 0.1cm},clip]{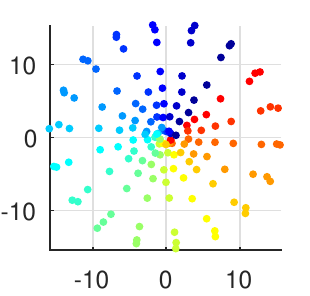}
			& \includegraphics[width = 0.19\textwidth, trim={0 0 0 0.1cm},clip]{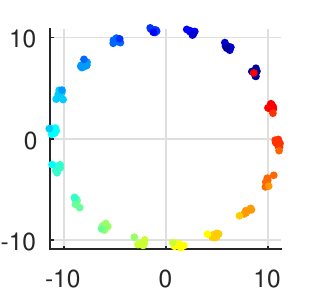}\\
			
			& {\small Fourth PC}
			& {\small Fourth PC}
			& {\small Fourth PC}
			& {\small Fourth PC}\\
			
	\end{tabular}
	
	\captionsetup{type = figure}
	\caption{The PCs obtained by projecting the multi-directional test set onto the PCA eigenvectors of the multi-directional test set. The top row shows the projections obtained using the 1\textsuperscript{st} and 2\textsuperscript{nd} eigenvectors of the multi-directional set. The bottom row shows the projections obtained using the 4\textsuperscript{th} and 5\textsuperscript{th} eigenvectors of the multi-directional set. The colour of each point shows the true sensor orientation (same colour bar as Fig.~\ref{fig:training}). }
	\label{tbl:visualtable}
	\vspace{-0.25cm}
\end{table*}

\begin{table*}[t]
	 \sffamily
	\centering

		\begin{tabular}{ccccc}

			& {\normalsize \textbf{Sliding Direction 0\degree{}}} & {\normalsize  \textbf{Sliding Direction 90\degree{}}} & {\normalsize  \textbf{Sliding Direction 180\degree{}}} & {\normalsize  \textbf{Motion Direction 270\degree{}}}\\

			\raisebox{0.5\totalheight}{\rotatebox[origin = lt]{90}{\mbox{\small Third PC} }}
			& \includegraphics[width = 0.19\textwidth ,  trim={0 0 0 0.1cm},clip]{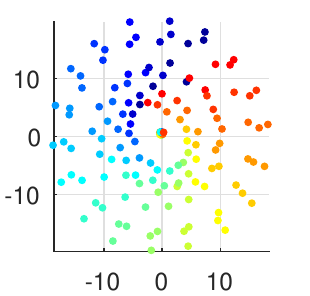}
			& \includegraphics[width = 0.19\textwidth, trim={0 0 0 0.1cm},clip]{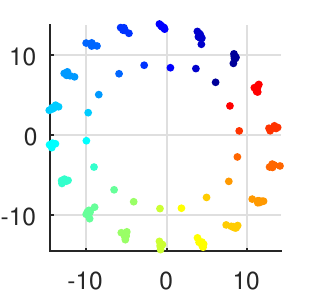}
			& \includegraphics[width = 0.19\textwidth, trim={0 0 0 0.1cm},clip]{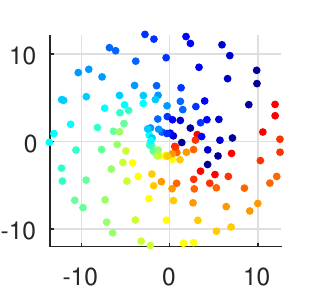}
			& \includegraphics[width = 0.19\textwidth, trim={0 0 0 0.1cm},clip]{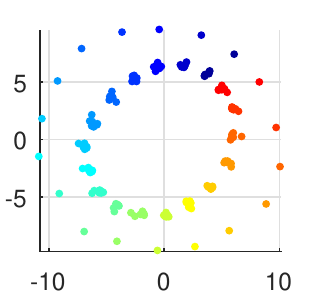}\\
			& {\small Second PC} 
			& {\small Second PC}
			& {\small Second PC}
			& {\small Second PC}\\

	\end{tabular}
	\captionsetup{type = figure}
	\caption{The PCs obtained by projecting the multi-directional test set onto  the 2\textsuperscript{nd} and 3\textsuperscript{rd} PCA eigenvectors of the training set (Fig.~\ref{fig:training}) . The colour of each point shows the true sensor orientation (same colour bar as Fig.~\ref{fig:training}). }
	\label{tbl:visualtableTrain}
	\vspace{-0.25cm}
\end{table*}
\subsubsection{Nonlinear Regression}
\label{sec:methodsGP}
We use nonlinear regression to map between the sensor data and the physical quantities of interest (the sensor lateral position and orientation). We use Gaussian Process Regression \cite{Rasmussen2005} with a Mat\'{e}rn 5/2 covariance function and a constant mean function. The GP is implemented using \cite{rasmussen2010} and the hyperparameters are found by optimising the log marginal likelihood \cite{Rasmussen2005} using a constrained optimiser (MATLAB optimisation toolbox).
 
Three independent GPs are used for regression: two to predict the sensor orientation, encoding the cosine (referred to as $\text{GP}_{\cos}$) and sine (referred to as $\text{GP}_{\sin}$) of the sensor orientation and a third for the sensor lateral position (referred to as $\text{GP}_{\text{lat}}$). The predicted sensor orientation is computed by using $\hat{\theta }= \atantwo(\bar{f}_{\sin\theta},\bar{f}_{\cos\theta})$, where $\bar{f}_{\sin\theta}$ is the mean prediction of $\text{GP}_{\sin}$ and $\bar{f}_{\cos\theta}$ is the mean prediction of $\text{GP}_{\cos}$. $\text{GP}_{\text{lat}}$ outputs the predicted lateral position $\hat{l }$. The spherical coordinates $\bm{v}$ are the inputs of the GPs and the outputs are $\hat{\theta }$ and $\hat{l }$.

We constructed baseline GPs to have a comparison method that does not use PCA. The baseline GPs perform regression using the raw sensor measurements directly using only one hyperparameter for all the pin positions. We use the baseline GPs to quantify the increase in perception accuracy obtained by including the PCA transform.

\subsection{Offline Testing data collection} \label{subsec: omni}

The offline test dataset is referred to as a multi-directional dataset as we collected the data over different sliding directions. We use the multi-directional set to evaluate the generalisation performance of the perception method which is trained using data recorded while performing taps. 

The multi-directional data collection consists of recording the tactile data whilst the sensor slides against the edge of a rectangular object (Fig.~\ref{fig:omnidirection}). The dataset comprises 1152 points (18 orientations $\times$ 8 sliding directions $\times$ 8 lateral positions), with 1 frame recorded at each sensor pose. 

At each sensor orientation (see Sec.~\ref{subsubsec:trainingProcedure}) the robot slides in different directions in 45\degree{} increments (Fig.~\ref{fig:omnidirection}). The starting position of each sliding movement is depicted by a black cross (Fig.~\ref{fig:omnidirection}). For each sliding direction, the sensor moves to 8 lateral positions in steps (refer to Sec.~\ref{sec:setup}), starting at position -6\,mm (on the surface) for motions from 90\degree{}--270\degree{}, and at position 9.9\,mm (off the surface) otherwise.

The sensor spends approximately the same amount of time in contact with the surface irrespective of the sliding direction. Therefore the sensor is only fully in contact with the surface for a length of 6\,mm along the sliding direction, except for the  90\degree{} and 270\degree{} motions where the sensor is in contact with the edge during all the lateral positions.

\subsection{Task: Contour Following}
\label{sec:methodsContour}

We achieve tactile exploration by combining the perception method with a simple controller (Fig.~\ref{fig:contourMethod}). The control policy \cite{Martinez-Hernandez2017} moves the sensor to the robot position $\bm{u}$ with $\Delta\bm{u}$ composed of: 1) 3\,mm linear exploratory movement $e$ along the perceived edge orientation and 2) a lateral correction movement perpendicular to the perceived edge orientation. We assume that the position of one point on the contour is known. The sensor orientation is fixed during the task and a proportional (P) controller with a hand tuned gain $K$ of 0.35 is used for the correction movement. The control policy (Fig.~\ref{fig:contourMethod}) is thus
\begin{equation}
\Delta\bm{u} = R(\hat{\theta})\begin{pmatrix} e \\ K( l_d-\hat{l})\end{pmatrix} 
\end{equation}
where $R(\theta)$ is the rotation matrix that aligns $e$ with the edge and $l_d$ is a constant representing the desired lateral position.

\begin{table*}[b]
	\vspace{-0.5em}
	\centering
	\caption{Summary of results.  Each column shows the RMS error obtained for each sliding direction which in turn is divided into ``On"  and ``Off" category. ``On" means the middle of the sensor is on the object surface and ``Off" means the middle of the sensor is off the surface.}
	\label{tbl:errortable}
	\setlength{\tabcolsep}{0.3em}
	\setlength{\extrarowheight}{0.4em}
	\begin{tabular}{cccccccccccccccccc}
		\hline
		& &  \multicolumn{2}{c}{\textbf{Direction 0\degree{}}}& \multicolumn{2}{c}{\textbf{Direction 45\degree{}}}& \multicolumn{2}{c}{\textbf{Direction 90\degree{}}}& \multicolumn{2}{c}{\textbf{Direction 135\degree{}}}& \multicolumn{2}{c}{\textbf{Direction 180\degree{}}}& \multicolumn{2}{c}{\textbf{Direction 225\degree{}}}& \multicolumn{2}{c}{\textbf{Direction 270\degree{}}}& \multicolumn{2}{c}{\textbf{Direction 315\degree{}}}\\
		\hline
		&\multicolumn{1}{c|}{} &On&\multicolumn{1}{c|}{Off}&On&\multicolumn{1}{c|}{Off}&\multicolumn{2}{c|}{On}&On&\multicolumn{1}{c|}{Off}&On&\multicolumn{1}{c|}{Off}&On&\multicolumn{1}{c|}{Off}&\multicolumn{2}{c|}{On}&On&Off\\
		GP baseline   &\multicolumn{1}{c|}{} &26.01\degree{}&\multicolumn{1}{c|}{16.59\degree{}}&2.94\degree{}&\multicolumn{1}{c|}{8.02\degree{}}&\multicolumn{2}{c|}{50.68\degree{}}&92.37\degree{}&\multicolumn{1}{c|}{99.59\degree{}}&127.18\degree{}&\multicolumn{1}{c|}{145.27\degree{}}&141.56\degree{}&\multicolumn{1}{c|}{153.39\degree{}}&\multicolumn{2}{c|}{90.32\degree{}}&50.10\degree{}&31.97\degree{}\\
		PCA + GP   &\multicolumn{1}{c|}{} &49.41\degree{}&\multicolumn{1}{c|}{10.17\degree{}}&24.17\degree{}&\multicolumn{1}{c|}{8.24\degree{}}&\multicolumn{2}{c|}{3.27\degree{}}&22.74\degree{}&\multicolumn{1}{c|}{10.82\degree{}}&46.74\degree{}&\multicolumn{1}{c|}{14.61\degree{}}&52.67\degree{}&\multicolumn{1}{c|}{11.98\degree{}}&\multicolumn{2}{c|}{12.66\degree{}}&51.87\degree{}&12.77\degree{}\\
		
		\hline
	\end{tabular}

\end{table*}
\section{RESULTS}

\subsection{Multi-directional Continuous Tactile Contact }\label{sec:resultOmni}

Here we examine the response of the sensor when sliding along the edge of a rigid object. We visualise the data using PCA to understand the sensor's response. Subsequently, the projections obtained using the tap training set are presented and lastly the accuracy of the perception method is evaluated.

\subsubsection{Multi-directional test data visualisation} \label{subsec:visualise}

PCA projections provide a low-dimensional vector representing the vector of tactile measurements $\bm{z}$. We plot this projection in 3D (see \cite{Aquilina2018}) to gain a better understanding of the underlying behaviour of the sensor with respect to the environment. We visualise the multi-directional set in Fig.~\ref{tbl:visualtable} to examine the effects of sliding motion; the colour shows the true sensor~orientation.
 
The 1\textsuperscript{st} and 2\textsuperscript{nd} PCs of the multi-directional set do not show a correlation between the sensor orientation and the PC1-PC2 angle ($\theta_{\text{PC12}}$), but instead, show a correlation to the sliding direction. This is shown in the first row of  Fig.~\ref{tbl:visualtable} by the location of the individual colours varying from panel to panel (left to right). Additionally, the magnitude of the vector comprising PC1 and PC2 is correlated to the magnitude of the shear, with a small magnitude for measurements which were recorded before the sensor started sliding. When the sensor slides along an object, shear force effects dominate the movement of the pins, which explains why sliding is represented in the first two PCs.

Higher PCs obtained from the multi-directional set represent sensor orientation features that are independent of the sliding direction. There is a large correlation between the angle of the projections in the PC4-PC5 plane and the true sensor orientation which is mostly invariant to the shear direction (bottom row of  Fig.~\ref{tbl:visualtable}). The 3\textsuperscript{rd} PC is not shown as it does not contain any information related to sensor~orientation.
\subsubsection{Multi-directional projections using the training set}
 \label{subsec:visualiseTrain}

The PCA-based perception algorithm uses a small training set (288 data frames) collected during discrete contacts with the object to infer orientation features in the continuous contact tactile dataset (1152 data frames). In  Fig.~\ref{tbl:visualtable} we visualised the multi-directional set using eigenvectors of sliding data. Here, we discuss a shear-invariant transformation, obtained by projecting the multi-directional dataset (Fig.~\ref{tbl:visualtableTrain}) onto the eigenvectors of the taps training set.

PC1 is correlated with the sensor pins expansion/compression which relates to the lateral position and not with the orientation features we are discussing here. 
  
PC2 and PC3 are correlated with the true sensor orientation (Fig.~\ref{tbl:visualtableTrain}). Thus, irrespective of the sliding direction, the tactile orientation features are still visible in the PC2-PC3 angle ($\theta_{\text{PC23}}$).

The same tactile features can be captured by using PCA eigenvectors of a discrete contact set without the need of a larger multi-directional set as shown by the similarity between the projections obtained using either set (compare the bottom row of  Fig.~\ref{tbl:visualtable} to  Fig.~\ref{tbl:visualtableTrain}). The eigenvectors of the multi-directional set (1152 data frames) capture orientation features better than the eigenvectors of the training set. This is expected since the dominant eigenvectors of the multi-directional set capture sliding; thus, the higher PCs are orthogonal to that sliding component. Nonetheless, the projections obtained using the eigenvectors of the training set are sufficiently good to represent a shear-invariant space with the advantage of a using much smaller dataset collected without sliding movements.

\subsubsection{Multi-directional offline test} \label{subsection:offline_errors}

Here we quantitatively analyse the effects of sliding motion on the sensor response to complement the qualitative analysis described above. The full perception method is considered in this section including the nonlinear regression that maps between the visualisations seen in  Fig.~\ref{tbl:visualtableTrain} to the sensor pose. The increase in performance obtained by using PCA prior to regression is evaluated by using the baseline GP model.

The root mean square (RMS) errors of the perceived sensor orientation ($\hat{\theta}$) are partitioned depending on the sliding direction and are further divided into two groups for each motion: ``On" denoting the middle of the sensor is on the object;  ``Off" denoting that the middle of the sensor is off the object. For the 90\degree{} and 270\degree{} motions, only the ``On" group is available since the middle of the sensor is always on the object surface.

Overall, the errors obtained for the ``Off" category are smaller than the ``On" category for both models (\autoref{tbl:errortable}). During the ``Off" motion the magnitude of the shear disturbance is smaller compared with the ``On" category explaining this result. This also agrees with the visualisations seen in  Fig.~\ref{tbl:visualtableTrain} and the bottom row of  Fig.~\ref{tbl:visualtable} where the correlation of the orientation feature is weaker in the centre of the plots where the magnitude of the shear disturbance is larger.

The errors obtained for the PCA-based perception algorithm are in most cases smaller than the baseline method. Importantly, the proposed model obtains a far better (7-15x) accuracy for the 90\degree{} and 270\degree{} motion, which is crucial as this is the main sliding direction used during contour-following. Therefore the projection performed by the PCA provides a good shear-invariant transform for this sliding direction.

Otherwise, for sliding motions starting from the outside of the object edge (direction 0\degree{}, 45\degree{} and 315\degree{}) the baseline model produces better results for the ``On" groups. In these directions, the main perturbation experienced by the sensor pins is a change in magnitude rather than direction thus enabling the baseline model to obtain the least errors in these directions. The PCA performs a transformation of the data which attenuates the influence of shear in some directions more than in others but which overall gives better accuracy in the majority of the sliding directions. This thus validates the use of PCA eigenvectors to project the pins into a low dimensional space less influenced by shear.

\begin{figure*}[t!]
			\vspace{0.15cm}
	\centering
	\begin{subfigure}{0.281\textwidth}
		\centering
		
		\begin{overpic}[width=\textwidth,trim = 0 0.07cm 0 0,clip]{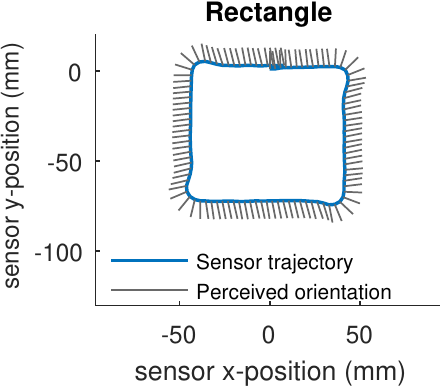}
			\put(10,80){\textbf{(a)}}
		\end{overpic}
		
	\end{subfigure}	
	\begin{subfigure}{0.22\textwidth}
		\centering
		
		\begin{overpic}[ width=\textwidth, trim = 0 0 0 0.01cm,clip]{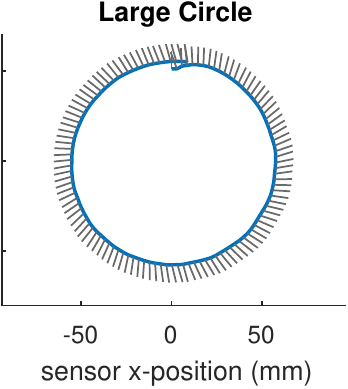}
			\put(-10,95){\textbf{(b)}}
		\end{overpic}
		
	\end{subfigure}	
	\begin{subfigure}{0.22\textwidth}
		\centering
		
		\begin{overpic}[width=\textwidth,trim = 0 0 0 0.01cm,clip]{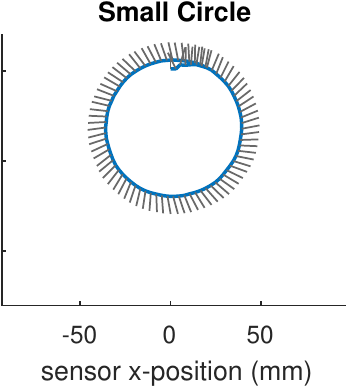}
			\put(-10,95){\textbf{(c)}}
		\end{overpic}
		
	\end{subfigure}	
	\begin{subfigure}{0.22\textwidth}
		\centering
		
		\begin{overpic}[width=\textwidth,trim = 0 0 0 0.01cm,clip]{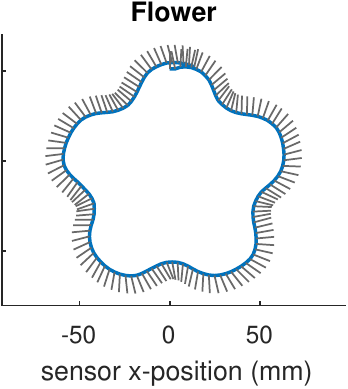}
			\put(-10,95){\textbf{(d)}}
		\end{overpic}
	\end{subfigure}

	\caption{(a) The trajectory (blue trace) performed by the robot when following the contour of a: (a) rectangle, (b) large circle, (c) small circle and (d) flower-like shape. The grey lines show the normal to the object edge as perceived by the sensor.}
	\label{fig:contours}
	\vspace{-1.2em}
\end{figure*}

\begin{figure}[h]
	\centering

\begin{subfigure}{0.255\textwidth}
	\centering
		\begin{overpic}[width=\textwidth,trim = 0 0 0 0.01cm,clip]{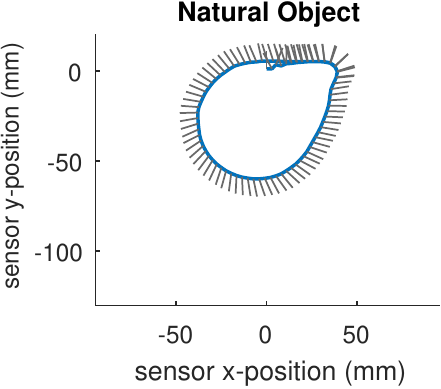}
		\put(70,25){\includegraphics[width=0.3\textwidth]{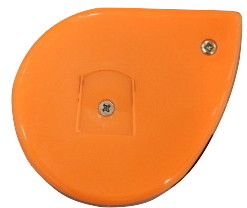}}
		\put(10,83){\textbf{(a)}}
	\end{overpic}

\end{subfigure}	
	\begin{subfigure}{0.2\textwidth}
	\centering

	\begin{overpic}[width=\textwidth,trim = 0.9cm 0 0 0.01cm,clip]{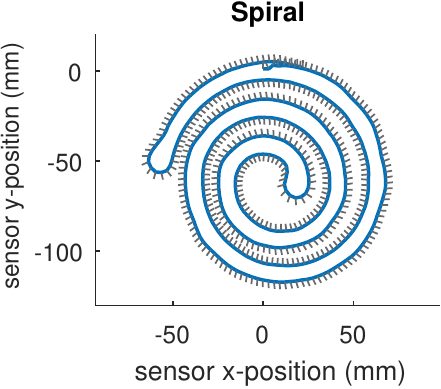}
		\put(5,93){\textbf{(b)}}
	\end{overpic}
	
\end{subfigure}	
	\caption{Natural object (tape measure) and spiral contour-following. The same colour coding of Fig.~\ref{fig:contours} is used here.}
	\label{fig:spiralContour}
\vspace{-1.6em}
\end{figure}
\subsection{Online Contour Following}

Finally, we present the online results for a tactile exploration task, specifically contour-following where the sensor continuously slides along the edge of various shapes.

We use four 3D-printed shapes, an acrylic laser-cut spiral shape, and a tape measure (Fig.~\ref{fig:setup}a) for this task. The shapes were chosen to test the limits of the method. The rectangle has zero curvature and corners. The two circles have different constant curvatures. The flower-like shape has both negative and positive curvatures. The tape measure is a natural object and has different curvature values and a corner. Finally, the spiral shape has negative and positive curvatures, contains corners and has closely spaced features.

Our approach successfully traced all contours (Fig.~\ref{fig:contours} and Fig.~\ref{fig:spiralContour}) by using the perception algorithm which consist of a PCA dimension reduction step (Sec.~\ref{sec:methodsPCA}) and nonlinear regression (Sec.~\ref{sec:methodsGP}) along with a simple control policy (Sec.~\ref{sec:methodsContour}). The RMS orientation errors are similar for all shapes with an average value of 12.2\degree{} which agrees with the errors obtained in Sec.~\ref{subsection:offline_errors} for the 270\degree{} sliding direction. The perception algorithm handles corners (Fig.~\ref{fig:contours}a and Fig.~\ref{fig:spiralContour}) by gradually perceiving a varying angle which changes smoothly over the corner. This was not a trained behaviour but emerged from the algorithm. The algorithm thus generalises to different curvatures.

The algorithm generalises from discrete contact tactile data collected on a zero curvature edge to data recorded whilst performing sliding motion over edges having varying curvature properties. The perception is not influenced by changes in curvature magnitude as shown by the two circles (Figs.~\ref{fig:contours}b, c) or by whether the curvature is positive or negative as demonstrated by the flower-like shape (Fig.~\ref{fig:contours}d). The method also performed contour-following of a natural object (Fig.~\ref{fig:spiralContour}a). The robustness of the perception is demonstrated with the spiral task (Fig.~\ref{fig:spiralContour}b) where apart from curvature variations, a different material is used from the training set and an upper limit on the lateral position perception error is imposed. The perceived lateral position errors do not impact the success of the contour-following as shown by the smooth shape-preserving trajectories achieved.

\FloatBarrier
\section{DISCUSSION}
The paper has demonstrated that a PCA-based perception algorithm can infer edge features invariant of the sliding motion. We achieved successful continuous contour-following of a wide range of shapes, thus showing that the simple features found by PCA with a suitable discrete contact training set are robust to sliding and to changes in curvature.

The PCA part of the algorithm also helps to visualise the sliding data which builds upon the tactile visualisation presented by Aquilina et al. \cite{Aquilina2018}. Here we show that visualisation of the multi-directional data reveals continuous tactile data is history dependent on the sliding direction. However, PCA can also encode the sliding vector in each sensor frame, which may be useful in experiments where sliding is a quantity of interest.

Additionally, due to the simple features extracted using PCA, the algorithm was able to generalise to various flat objects of an unknown shape despite being trained only on a portion of a straight edge. This is related to the work by Luo et al. \cite{Luo2014} where a specific data descriptor yields rotation and translation invariance. However, the most common approach is to obtain invariance by training on samples which have different properties such as the shape-invariant method presented by Yuan et al. \cite{Yuan2017}. This would have been equivalent to using a sliding contact training set in this work. However, in practice it may not be possible to train for every event the system could encounter; instead we view it desirable to have a robust system that produces reasonable results on a wide range of scenarios.

A limitation of this work is that it only considers planar 2D objects which constrains the tactile exploration problem considerably. However, this still demonstrates the proof of concept. It would be instructive to test the method with objects that do not have clear cut edges. Another limitation of the current method is that the lateral displacement predictions are not as accurate (overall error of 2.15\,mm for the offline task) as the orientation predictions. These limitations will be addressed in future work.

We expect the proposed method to apply to more challenging tasks due to its generalisation capabilities. Such a method could be combined with more complicated controllers using predictive or adaptive control to complete demanding manipulation or exploration tasks in unstructured environments.


\bibliographystyle{IEEEtran}

\end{document}